\pdfoutput=1

\documentclass[11pt]{article}

\usepackage{acl}
\usepackage{textgreek}
\usepackage{times}
\usepackage{latexsym}
\usepackage{mathtools}
\usepackage{amsmath}
\usepackage{txfonts}
\usepackage{xspace}
\usepackage{multirow}
\usepackage{enumitem} 
\newcommand{\macrof}{$\mathrm{m}$-$\mathrm{F_1}$\xspace}
\newcommand{\microf}{$\mathrm{\muup}$-$\mathrm{F_1}$\xspace}

\newcommand{\cls}{\texttt{\small [cls]}\xspace}
\newcommand{\sep}{\texttt{\small [sep]}\xspace}

\usepackage{microtype}

\title{Processing Long Legal Documents with Pre-trained Transformers: Modding LegalBERT and Longformer}

\author{
  Dimitris Mamakas\thanks{\hspace{0.5em}Equal contribution.}$^{\;\;\;\dagger\;}$ \quad
  Petros Tsotsi$^{\ast\;\dagger\;}$ \\
  \bf Ion Androutsopoulos$^{\;\dagger\;}$ \quad
  Ilias Chalkidis$^{\;\;\ddagger\;\diamond\;}$ \\
$^{\dagger\;}$ Department of Informatics, Athens University of Economics and Business, Greece\\
${\;\ddagger}$ Department of Computer Science, University of Copenhagen, Denmark \\
$^{\diamond\;}$ Cognitiv+, Athens, Greece\\
}

\date{}

\begin{document}
    \maketitle

\begin{abstract}
Pre-trained Transformers currently dominate most NLP tasks. They impose, however, limits on the maximum input length (512 sub-words in BERT), which are  too restrictive in the legal domain. Even sparse-attention models, such as Longformer and BigBird, which increase the maximum input length to 4,096 sub-words, severely truncate texts in three of the six datasets of LexGLUE. Simpler linear classifiers with TF-IDF features can handle texts of any length, require far less resources to train and deploy, but are usually outperformed by pre-trained Transformers. We explore two directions to cope with long legal texts: (i) modifying a Longformer warm-started from LegalBERT to handle even longer texts (up to 8,192 sub-words), and (ii) modifying LegalBERT to use TF-IDF representations. The first approach is the best in terms of performance, surpassing a hierarchical version of LegalBERT, which was the previous state of the art in LexGLUE. The second approach leads to computationally more efficient models at the expense of lower performance, but the resulting models still outperform overall a linear SVM with TF-IDF features in long legal document classification.
\end{abstract}

\section{Introduction}

Transformer-based models \cite{Vaswani2017}, like BERT \cite{Bert}, RoBERTa \cite{liu-2019-roberta}, and their numerous offspring, currently dominate most natural language processing (NLP) tasks. These models are pre-trained on very large corpora using generic tasks (e.g., masked token prediction) that do not require human annotations, and are then fine-tuned (further trained) on typically much smaller task-specific datasets with manually annotated ground truth. The quadratic complexity of their attention mechanisms, however, imposes limits on the maximum input length (512 sub-word tokens in BERT, RoBERTa), which are often too restrictive in the legal domain, where longer documents are common. The same restrictions apply to LegalBERT \cite{chalkidis-etal-2020-legalbert}, a BERT variant pre-trained on legal corpora.

\begin{figure*}
    \centering
    \resizebox{\textwidth}{!}{
    \includegraphics{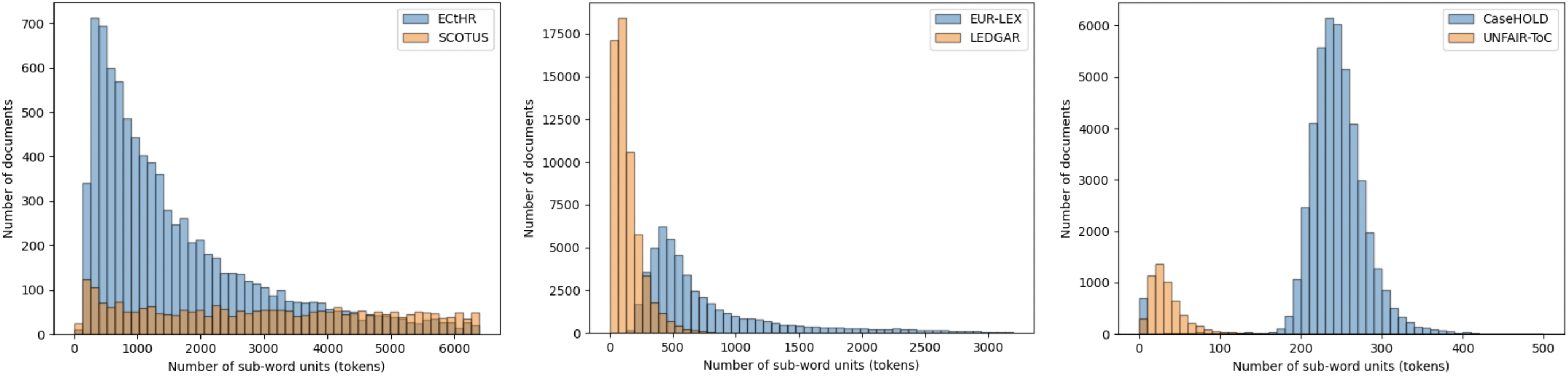}
    }
    \vspace{-4mm}
    \caption{Distribution of input text length, measured in BERT sub-word tokens, across the six LexGLUE datasets. Copied with permission from \citet{chalkidis-etal-2021-lexglue}.}
    \label{fig:lengths}
    \vspace{-2mm}
\end{figure*}

Even the sparse-attention Longformer \cite{Longformer}, a well-known Transformer that increases the maximum input to 4,096 sub-words, severely truncates texts in three of the six datasets (see Fig.~\ref{fig:lengths}) of the LexGLUE legal NLP benchmark \cite{chalkidis-etal-2021-lexglue}. On the other hand, simpler linear classifiers with TF-IDF features \cite{ManningIR}, which were very common before deep learning, can handle texts of any length, at least in text classification tasks, require far less resources to train and deploy, but are nowadays usually outperformed by pre-trained Transformers. 

\begin{figure}[t]
    \centering
    \resizebox{\columnwidth}{!}{
    \includegraphics{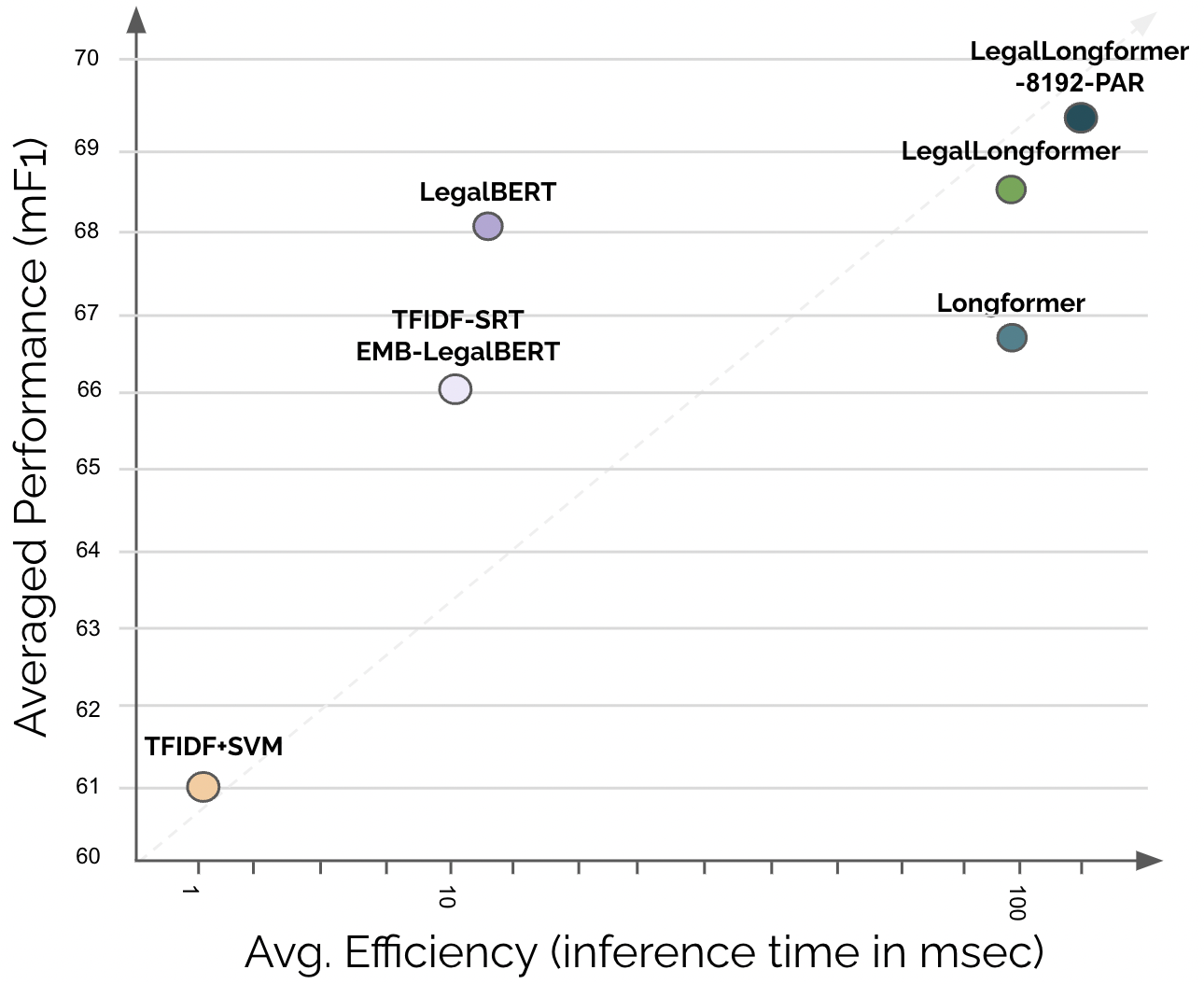}
    }
    \caption{Comparison of examined models presented in Section~\ref{sec:examined_models}, considering averaged down-stream performance and efficiency (inference time) in LexGLUE long document classification tasks (ECtHR, SCOTUS).}
    \label{fig:efficiency}
    \vspace{-4mm}
\end{figure}

Motivated by these observations, we explore two directions to better cope with long legal texts: (i) we modify a Longformer  warm-started from LegalBERT to handle even longer texts (up to 8,192 sub-words), a resource-intensive direction that further increases the parameters and processing time of large sparse-attention Transformer models; and (ii) we modify LegalBERT to use TF-IDF representations, which allows processing longer texts without increasing the model sizes. The first approach is the best overall in terms of performance, surpassing a hierarchical version of LegalBERT \cite{chalkidis-etal-2021-paragraph}, which was the previous state of the art in LexGLUE. The second direction leads to computationally more efficient models at the expense of lower performance, but still outperforms overall a linear Support Vector Machine (SVM) \cite{SVM} with TF-IDF features in long document classification.

\section{Related Work} \label{sec:RelatedWork}

\subsection{Long Document Processing}
Transformer-based models consist of stacked Transformer blocks \cite{Vaswani2017}. Each block builds a revised embedding (vector representation) for each (sub-word) token of the input text, based on the embeddings of the previous block, starting from an initial embedding layer that provides an embedding per vocabulary token. For a block with a single attention head and an input $n$ tokens long, generating a single revised token embedding involves computing a weighted sum (weighted by attention scores) over the $n$ token embeddings of the previous block. Hence, $O(n^2)$ time is required to generate all the $n$ revised token embeddings. With $k$ attention heads, the complexity is $O(k \cdot n^2)$.\vspace{2mm}

\noindent \textbf{Sparse-attention} variants of Transformers, like those used in Longformer \cite{Longformer}, ETC \cite{ainslie-etal-2020-etc}, BigBird \cite{BigBird}, generate each revised token embedding by attending (considering) only the previous block's embeddings for the current, the $l$ previous, and the $l$ next tokens in the input text, i.e., the weighted sum is now over only $2 \cdot l+1$ (equal to 512 by default) token embeddings of the previous block. The complexity becomes $O(k \cdot n \cdot l)$, linear to $n$. To better capture long-distance dependencies, these models also use \emph{global attention}. This involves either standard pseudo-tokens, such as the \cls token at the beginning of each text, or additional pseudo-tokens, e.g., \sep tokens placed at the end of each paragraph. In both cases, these special global tokens are attended by, and attend all other tokens, allowing information to flow across distant tokens, even when sparse attention is used. 

We experiment with Longformer, a well-known and relatively simple sparse-attention Transformer, which can process texts up to 4,096 sub-words long. ETC and Big Bird use the same maximum input length and are very similar; one difference is that they employ additional pre-training objectives for the global tokens, whereas in Longformer the global tokens are not pre-trained. We also expand Longformer to process texts up to 8,192 sub-words long, and we consider an ETC-like global attention scheme with additional \sep tokens.\vspace{2mm}

\noindent \textbf{Hierarchical} Transformers, e.g., SMITH \cite{SMITH}, use BERT (or other base models) to separately encode each paragraph or other segments of the input text that do not exceed the base model's maximum input length. The generated paragraph embeddings (e.g., the embeddings of \cls tokens placed at the beginning of each paragraph) are then passed through additional stacked Transformer blocks, to allow interactions between the paragraph embeddings. The resulting context-aware paragraph embeddings can then be used to classify individual paragraphs or to classify the entire text (e.g., using the first paragraph embedding or by max-pooling over all paragraph embeddings). 

In LexGLUE \cite{chalkidis-etal-2021-lexglue} a similar hierarchical model \cite{chalkidis-etal-2021-paragraph} was used, with either generic BERT variants (e.g., BERT, RoBERTa, DeBERTa) or LegalBERT as the base model, in three of the benchmark's tasks (ECtHR Task A and B, SCOTUS) where the average text length was much higher than BERT's maximum length (Fig.~\ref{fig:lengths}). Unlike SMITH, the additional paragraph-level Transformer blocks were not pre-trained. We compare against this hierarchical variant of LegalBERT on LexGLUE.\vspace{2mm}

\noindent \textbf{Recurrent} Transformers are another approach to handle long texts \cite{TransformerXL,XLNET,ERNIEDOC}. We do not consider them here due to the latency that recurrency introduces. \vspace{2mm}

\noindent \textbf{Bag-of-Word} (BoW) models typically represent each text as a (sparse) feature vector $\left<f_1, \dots, f_{|V|}\right>$, with one feature $f_i$ per vocabulary word. TF-IDF features \cite{ManningIR} are common. Given a text $n$ words long, each feature $f_i$ becomes:
\[ f_i = \textit{TF}_i \cdot \textit{IDF}_i = \frac{c_i}{n} \cdot \log \frac{N}{1 + d_i}, \]
where $c_i$ is the frequency of the $i$-th vocabulary word in the text, $N$ is the number of documents in a corpus (in text classification this is often the training set), and $d_i$ counts the documents of the corpus that contain the $i$-th vocabulary word.\footnote{In `sublinear' TF-IDF, a logarithm is also applied to TF.} Averaging word embeddings \cite{peng-etal-2016,brokos-etal-2016-using} with or without TF-IDF weighting, is also a BoW representation, but typically performs worse, since averaging leads to very noisy representations.

Such BoW representations discard word order, but are also insensitive to the length of the input text, in the sense that the feature vector always contains $|V|$ features. Combining TF-IDF feature vectors with linear classifiers leads to models that can handle texts of any length and require far less resources to train compared to modern Transformer-based models, at the expense of lower performance.\vspace{2mm}

\noindent \textbf{BoW-BERT}:
Our attempts to combine TF-IDF features with BERT were inspired by the work of \citet{hessel-BoW-BERT}, who reported that shuffling the words of each text during fine-tuning led to a degradation of less than 5 p.p.\ (F1 or accuracy) of BERT's performance in most GLUE tasks \cite{GLUE}.\footnote{However, \citet{hessel-BoW-BERT} also cite other work that found word shuffling to have a larger impact on pre-trained Transformers (and LSTMs) in other datasets and tasks. \citet{sinha-etal-2021-masked} and \citet{abdou-etal-2022-word} investigated the effect of word shuffling on pre-trained Transformers in more detail, considering mostly word shuffling during pre-training.} The resulting model, called BoW-BERT, can be seen as operating on BoW representations, in the sense that word order is lost. \citet{hessel-BoW-BERT} also reported that BoW-BERT performed better than other BoW models on GLUE, including linear models with TF-IDF features. 
BoW-BERT's word shuffling, however, does not change the text length, hence it does not address BERT's maximum input length limit; IDF information is also not considered. 

By contrast, we remove multiple occurrences of the same word from each text; to incorporate TF-IDF information, we order the remaining words by TF-IDF and/or we add a TF-IDF embedding layer, both discussed below.

\subsection{Applications in Legal NLP}

In the early days of Deep Learning for legal NLP, the community examined the use of the Hierarchical Attention Network (HAN) of \citet{yang-etal-2016-hierarchical} or simpler variants (hierarchical BILSTMs) to encode long documents in applications of legal judgment prediction for Chinece \cite{zhong-etal-2018-legal} or ECtHR \cite{chalkidis-etal-2019-neural} court cases, showcasing improvement over flat RNN-based models, such as stacked BILSTMs followed by a single-head attention layer \cite{pmlr-v37-xuc15}. Hierarchical BILSTMs with self-attention were also employed by \citet{chalkidis-etal-2018-obligation} for sequential sentence classification in order to identify obligations and prohibitions in contractual paragraphs.

Hierarchical variants of Transformers were initially proposed by \citet{chalkidis-etal-2019-neural}. In their work, document paragraphs are encoded via a shared BERT encoder to produce paragraph embeddings, which are then combined with max-pooling to form the final document embedding. This model outperformed strong RNN-based methods such as the Hierarchical Attention Network (HAN). 

Later on, \citet{chalkidis-etal-2021-paragraph} presented a new variant, where the paragraph embeddings are fed into additional stacked Transformer blocks, to allow cross-paragraph contextualization.
This latter version has also been used in other legal NLP applications, by \citet{niklaus-etal-2021-swiss, niklaus-etal-2022-cross} in judgment prediction of Swiss court cases, using XLM-R as the underlying encoder, and by \citet{chalkidis-etal-2021-lexglue} for the long document classification tasks of LexGLUE, using several alternative pre-trained Transformers, alongside Longformer. \citet{XIAO202179} released a Longformer pre-trained on Chinese legal corpora, which outperforms baselines in several legal NLP tasks. More recently, \citet{dai-et-al-2022-hierarchical} explored how tunable hyper-parameters of Hierarchical Transformers and Longformer, such as the size of the local window, affect downstream performance. In experiments on the ECtHR dataset, they found that fewer but larger local windows (paragraphs), e.g., 8$\times$512, instead of 32$\times$128, in Hierarchical Transformers improve performance. 

Hierarchical Transformers are also used in the work of \citet{malik-etal-2021-ildc} in legal judgment prediction of Indian court cases, where their best-performing model uses XLNet \cite{XLNET} as the underlying paragraph encoder followed by stacked BiGRUs. Moreover, hierarchical Transformers similar to those of \citeauthor{malik-etal-2021-ildc} have been also used by \citet{kalamkar-etal-2022-corpus} for sequential legal sentence classification in order to segment Indian court cases into topical and coherent parts.

\section{Models Considered}
\label{sec:examined_models}

We discuss models \citet{chalkidis-etal-2021-lexglue} evaluated on LexGLUE as baselines, and models we introduce. The LexGLUE baselines also included RoBERTa \cite{liu-2019-roberta}, DeBERTa \cite{he2021deberta}, BigBird \cite{BigBird}, and CaseLaw-BERT \cite{Zheng2021}, which are not considered here. \citet{chalkidis-etal-2021-lexglue} found RoBERTa and DeBERTa to be better than BERT on LexGLUE, but worse than LegalBERT; no legally pre-trained variants of RoBERTa and DeBERTa are available. BigBird and CaseLaw-BERT were found to be overall slightly worse than Longformer and LegalBERT, respectively, on LexGLUE.

\subsection{LexGLUE baselines} \label{sec:LexGLUEbaselines}

\noindent\textbf{TFIDF-SVM} is a linear SVM with TF-IDF features for the top-$K$ most frequent word $n$-grams of the training set, where $n\in[1,2,3]$.\footnote{$K\in[20k, 30k, 40k]$ is tuned per task on dev.\ data.}\vspace{2mm}

\noindent\textbf{LegalBERT} \cite{chalkidis-etal-2020-legalbert} is BERT pre-trained on English legal corpora (legislation, contracts, court cases). In the long document classification tasks (see Table~\ref{tab:datasets_summary}), we deploy its hierarchical variant (Section~\ref{sec:RelatedWork}) as in \citet{chalkidis-etal-2021-lexglue}.\vspace{2mm}

\noindent\textbf{Longformer} \cite{Longformer}. This is the original Longformer, discussed in Section \ref{sec:RelatedWork}. It extends the maximum input length to 4,096 sub-word tokens. Like BERT and RoBERTa, Longformer uses absolute positional embeddings, i.e., there is a separate positional embedding for each token position up to the maximum input length. Longformer's positional embeddings were warm-started from the 512 positional embeddings of RoBERTa, cloning them 8 times (e.g., the embeddings of positions 513--1024 were initialized to the same RoBERTa positional embeddings as positions 1--512). All the other parameters of Longformer (and RoBERTa) are not sensitive to token positions and were warm-started from the corresponding RoBERTa parameters.\footnote{E.g., the dense layers that produce the attention's query, key, value embeddings are the same for all token positions.} After warm-starting, Longformer was further pre-trained for 64k steps on generic corpora.

\subsection{Extensions of LegalBERT}

\noindent\textbf{TFIDF-SRT-LegalBERT}:
This is LegalBERT, but we remove duplicate sub-words from the input text and sort the remaining ones by decreasing TF-IDF during fine-tuning. Removing duplicate words is an attempt to avoid exceeding the maximum input length. In ECtHR, for example, the average text length (in sub-words) drops from 1,619 to 1,120; in SCOTUS, from 5,953 to 1,636 (see Fig.~\ref{tab:datasets_summary}). If the new form of the text still exceeds the maximum input length, we truncate it (keeping the first 512 tokens). Ordering sub-words by decreasing TF-IDF hopefully allows the model to learn to attend earlier sub-words (higher TF-IDF) more, utilizing BERT's positional embeddings as TF-IDF ranking encodings. This is a BoW model, since the original word order of the input text is lost.\vspace{2mm}

\begin{figure*}[t]
\resizebox{\textwidth}{!}{
\includegraphics{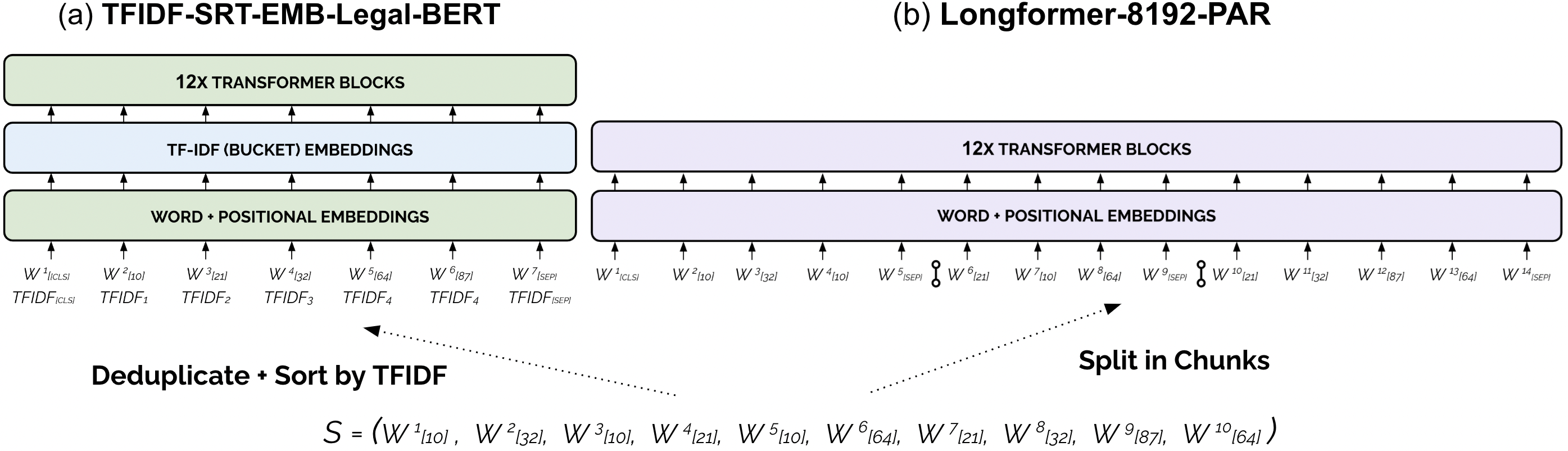}
}
\caption{(a) TFIDF-SRT-EMB-LegalBERT, with sub-word token deduplication, re-ordering by TF-IDF, and TF-IDF embedding layer. (b) Longformer-8192-PAR extended to encode up to 8192 sub-word tokens, split into paragraphs separated by \sep tokens. The original sequence ($S$) of sub-words ($W$) is shown in the bottom. Superscripts ($W^p$) denote positioning in each sequence. Subscripts ($W_{[id]}$) are the indices of the sub-words in the model's vocabulary. In both models, the resulting contextualized \cls token embedding is fed to a linear classifier.}
\label{fig:models}
\end{figure*}

\noindent\textbf{TFIDF-SRT-EMB-LegalBERT}:
The same as the previous model, except that we add a TF-IDF embedding layer (Fig.~\ref{fig:models}). We bucketize the distribution of TF-IDF scores of the training set and assign a TF-IDF embedding to each bucket. During fine-tuning, we compute the TF-IDF score of each sub-word (before deduplication) and we add the corresponding TF-IDF bucket embedding to each token's input embedding when its positional embedding is also added. The TF-IDF bucket embeddings are initialized randomly and trained during fine-tuning. 
Hence, this model is informed both about TF-IDF token \emph{ranking} (via word re-ordering) and TF-IDF \emph{scores} (captured by TF-IDF embeddings). This is still a BoW model, since it ignores the original word order, like the previous model.\vspace{2mm}

\noindent\textbf{TFIDF-EMB-LegalBERT}:
The same as LegalBERT, but we add the TF-IDF layer of the previous model. Token deduplication and ordering by TF-IDF scores are not included. This allows us to study the contribution of the TF-IDF layer on its own by comparing to the original LegalBERT. The resulting model is aware of word-order via its positional embeddings (like BERT and LegalBERT). For long texts, it addresses the maximum input length limitation via its hierarchical variant, which is similar to LegalBERT's \cite{chalkidis-etal-2021-lexglue}. 

\subsection{Extensions of Longformer}

\noindent\textbf{Longformer-8192}: This is the same as the original Longformer \cite{Longformer}, which was warm-started from RoBERTa (Section~\ref{sec:LexGLUEbaselines}), but we extend the maximum input length to 8,192 sub-words. We warm-start the positional embeddings from those of Longformer, cloning them once (positions 4,097--8,192 get the same initial embeddings as positions 1--4,096). To keep the computational complexity under control, we decrease the local attention window size from 512 to 128 sub-words.\footnote{Table~\ref{tab:efficiency_scores} shows that despite this counter-measure, the expansion to 8,192 sub-words leads to almost 2$\times$ inference time and 30\% increase in memory.} 
All parameters, including positional embeddings, are updated during fine-tuning, again as in the original Longformer. We did not perform any additional pre-training, however, beyond that of the original Longformer, lacking computing resources. All Longformer variants are aware of word order.\vspace{4mm}

\renewcommand{\arraystretch}{1.2}
\begin{table*}[t]
\centering
\resizebox{\textwidth}{!}{
\begin{tabular}{l|c|c|c|ccc|c}
\multirow{2}{*}{\textbf{Dataset}}        & \multirow{2}{*}{\textbf{Source}} & \multicolumn{2}{c|}{\bf Text Length (Original/Unique)} & \multicolumn{3}{c|}{\textbf{Instances}} & \multirow{2}{*}{\textbf{Classes}}  \\ 
& & Average & Maximum  & Train & Dev. & Test & \\
\hline
\multicolumn{8}{c}{\textsc{Long Document Classification Tasks}} \\
\hline
ECtHR (Task A) & \cite{chalkidis-etal-2019-neural}      & 1.6k (1.1k) & 35.4k (27.2k) & 9,000 & 1,000 & 1,000     & 10+1      \\
ECtHR (Task B) & \cite{chalkidis-et-al-2021-ecthr}     & 1.6k (1.1k) & 35.4k (27.2k)  & 9,000 & 1,000 & 1,000     & 10+1      \\
SCOTUS         & \cite{spaeth2020}                       & 6.0k (1.6k) & 88.6k (12.1k)  & 5,000 & 1,400 & 1,400     & 14        \\
\hline
\multicolumn{8}{c}{\textsc{Short Document Classification Tasks}} \\
\hline
EUR-LEX$^\ast$        & \cite{chalkidis-etal-2021-paragraph}  & 1.1k (341) & 140.1k (10k)  & 55,000 & 5,000 & 5,000    & 100       \\
LEDGAR         & \cite{tuggener-etal-2020-ledgar}        & 113 (65) & 1.2k (484)      & 60,000 & 10,000 & 10,000  & 100       \\
UNFAIR-ToS     & \cite{lippi-etal-2019-claudette}        & 33 (25) & 441 (181)         & 5,532 & 2,275 & 1,607     & 8+1       \\
\end{tabular}
}
\caption{LexGLUE statistics. Lengths in sub-word tokens, before and after (in brackets) deduplication. $^\ast$EUR-LEX is treated as a short document task in our work, since using only the first 512 tokens (what we do) has comparable performance with using the full texts~\cite{chalkidis-etal-2019-large}. +1 denotes an extra class for no-label instances.}
\label{tab:datasets_summary}
\end{table*}

\noindent\textbf{Longformer-8192-PAR}: This is the same as the previous model, but we place a global token (Section~\ref{sec:RelatedWork}), specifically a \sep token, at the end of each paragraph (Fig.~\ref{fig:models}). By contrast, the original Longformer and Longformer-8192 use the single \cls token at the beginning of the input text as a single global token for classification tasks.\footnote{Additional global tokens were used by \citet{Longformer} in other tasks, e.g., question answering.} As in the previous model, we decrease the local attention window size from 512 to 128 sub-words. 

Our intuition was that using more global tokens, and synchronizing them with paragraph breaks would allow information to flow more easily across paragraphs, viewed as discourse segments. Previous work by \citet{BigBird} also suggests that such ETC-like global attention layouts lead to better results.  Again, all parameters are updated during fine-tuning, but we did not perform any additional pre-training to better adjust the model to the new global attention layout, lacking resources.\vspace{2mm}

\noindent\textbf{LegalLongformer}: Similar to Longformer, but warm-started from LegalBERT. We clone the  positional embeddings of LegalBERT eight times to cover positions 1--4,096 (instead of 1--512 in LegalBERT) and update them during fine-tuning. All other parameters are also warm-started from LegalBERT and are updated during fine-tuning. Following \citet{Longformer}, we warm-start the global attention parameters of LegalLongformer with the (local) attention parameters of LegalBERT. Again, no additional pre-training was performed.\vspace{2mm}

\noindent\textbf{LegalLongformer-8192}: Similar to Longformer-8192, but again warm-started from LegalBERT. In this case, we clone the  positional embeddings of LegalBERT 16 times to cover positions 1--8,192. Again, no additional pre-training was performed.\vspace{2mm}

\noindent\textbf{LegalLongformer-8192-PAR}: The same as the previous model, but with global tokens at the end of each paragraph, as in Longformer-8192-PAR.

\section{Experiments} 

\subsection{Datasets}

LexGLUE \cite{chalkidis-etal-2021-lexglue} is a collection of six simplified English legal NLP datasets that are used to evaluate the performance of NLP methods across seven legal text understanding tasks. Inspired by GLUE \cite{GLUE} and SuperGLUE \cite{wang-2019-superglue}, LexGLUE was designed to push towards generic-pretrained models that can cope with multiple legal NLP tasks with limited extra training (fine tuning) for each one. 

Here, we experiment with six of the seven tasks of LexGLUE, excluding CaseHOLD \cite{Zheng2021}, a multiple choice question answering task about holdings of US court cases. The other six tasks are all framed as text classification problems. While our work targets the long document classification tasks (ECtHR Tasks A and B, SCOTUS), we also experiment with tasks that involve short texts (EUR-LEX, LEDGAR, UNFAIR-ToS), for completeness. Table~\ref{tab:datasets_summary} lists the sources of the datasets we experiment with and provides key statistics. ECtHR Task A and B require deciding which articles of the European Convention of Human Rights were violated, or allegedly violated, respectively; both tasks use the same dataset in LexGLUE. SCOTUS requires classifying opinions of the US Supreme Court into issue areas (e.g., Criminal Procedure, Civil Rights). EUR-LEX requires labeling European laws with concepts from a European Union taxonomy. LEDGAR requires assigning topical categories to contract provisions. UNFAIR-ToS requires detecting unfair terms in terms of service. Consult \citet{chalkidis-etal-2021-lexglue} and the work cited in Table~\ref{tab:datasets_summary} for further information.

\subsection{Evaluation measures}
Following \citet{chalkidis-etal-2021-lexglue}, for each task we report 
macro-F1 (\macrof), which assigns equal importance to all classes, and micro-F1 (\microf), which assigns more importance to frequent classes. 

\subsection{Experimental setup}
Across all experiments, we use Adam \cite{Kingma2015} with initial learning rate 3e-5. We train models up to 20 epochs using early stopping, monitoring \microf on the development data. We run all experiments with 5 different random seeds and report test results for the seeds with the best development scores. For the TF-IDF bucket embedding layer, we search in \{16, 32, 64, 128\} for the number of buckets that maximizes \microf on the development data, separately for each task.

\begin{table*}[t]
    \centering
    \resizebox{\textwidth}{!}{
    \begin{tabular}{l|cc|cc|cc|cc|cc|cc|c|c}
         \multirow{2}{*}{\bf Method}  & \multicolumn{2}{c|}{\bf ECtHR (A)*} & \multicolumn{2}{c|}{\bf ECtHR (B)*} & \multicolumn{2}{c|}{\bf SCOTUS*} & \multicolumn{2}{c|}{\bf EUR-LEX} & \multicolumn{2}{c|}{\bf LEDGAR}  & \multicolumn{2}{c|}{\bf UNFAIR-ToS} \\
         & \microf & \macrof & \microf & \macrof & \microf & \macrof & \microf & \macrof & \microf & \macrof & \microf & \macrof \\
         \hline
         \hline
         \multicolumn{13}{c|}{BoW models (word order lost)} \\
         \hline
         \hline
         TFIDF-SVM                      & 62.6 & 48.9 & 73.0 & 63.8 & \underline{74.0} & \underline{64.4} & 63.4 & 47.9  & \underline{87.0} & \underline{81.4} & 94.7 & 75.0 \\
         \hline
        TFIDF-SRT-LegalBERT             & \underline{69.8} & 62.8 & 78.5 & 71.9 & 73.4 & 61.8 & 69.6 & 53.7 & 86.9 & 80.8 & 95.3 & \underline{80.6} \\
        TFIDF-SRT-EMB-LegalBERT          & 68.7 & \underline{63.1} & \underline{79.0} & \underline{72.5} & 73.9 & 63.6 & \underline{69.7} & \underline{53.9} & 86.5 & 80.3 & \underline{95.8} & 78.7 \\
         \hline
         \hline
         \multicolumn{13}{c|}{LegalBERT variants that retain word order} \\
         \hline
         \hline
        LegalBERT   & \underline{70.0} & \underline{64.0} & \underline{80.4} & \underline{74.7} & \underline{76.4} & \underline{66.5} & \underline{72.1} & \underline{57.4} & 88.2 & 83.0 & \bf \underline{96.0} & \bf \underline{83.0} \\
         TFIDF-EMB-LegalBERT        & \underline{70.0} & 61.9 & 79.4 & 73.5 & 74.9 & 64.7 & 71.6 & 56.9 &  \underline{88.7} &  \underline{83.4} & 95.9 & 82.1 \\
        \hline
         \hline
         \multicolumn{13}{c|}{Longformer variants (all retain word order)} \\
         \hline
         \hline
         Longformer   & 69.9 & 64.7 & 79.4 & 71.7 & 72.9 & 64.0 & 71.6 & \bf\underline{57.7} & 88.2 & 83.0 & 95.5 & \underline{80.9} \\
Longformer-8192                & 70.9 & 62.1 & 79.2 & 73.9 & 73.7 & 63.6 & \multicolumn{6}{c|}{(Not considered for short-document tasks.)} \\
Longformer-8192-PAR                 & 70.8 & 62.3 & 79.0 & 73.1 & 73.9 & \underline{66.0}  & \multicolumn{6}{c|}{$>>$} \\
        \hline
    LegalLongformer                & \underline{\bf 71.7} & 63.6 & 80.5 & \bf \underline{76.4} & 76.6 & 66.9 & \underline{72.2} & 56.5 & \bf \underline{88.8} & \bf \underline{83.5} & \underline{95.7} & 80.6 \\
LegalLongformer-8192              & 71.2 & 64.3 & \bf \underline{81.4} & 74.2 & \bf \underline{77.5} & \bf \underline{67.3} & \multicolumn{6}{c|}{(Not considered for short-document tasks.)} \\
LegalLongformer-8192-PAR              & 71.4 & \bf \underline{68.4} & 79.6 & 73.9 & 76.2 & 66.3 & \multicolumn{6}{c|}{$>>$} \\
    \end{tabular}
    }
    \caption{Test results across LexGLUE tasks considered. In starred tasks, we use the hierarchical variant of LegalBERT. We do not consider  extended Longformers in short document classification tasks (last three; see also Table~\ref{tab:datasets_summary}), which are included for completeness. Best scores per group are \underline{underlined}, and best overall are in \textbf{bold}.
    }
    \label{tab:leaderboard}
\end{table*}

\subsection{Experimental results}

Table~\ref{tab:leaderboard} lists the test results of all models across the six tasks considered. Table~\ref{tab:avg_long_leaderboard} aggregates the test results over the three long-document classification tasks (ECtHR Tasks A and B, SCOTUS) we are mainly interested in (see also see Table~\ref{tab:datasets_summary}). We use the harmonic mean over the scores of the three tasks, following \citet{shavrina2021how}.\vspace{4mm}

\noindent\textbf{BoW models}: The results of the two BoW variants of LegalBERT (TFIDF-SRT-LegalBERT, TFIDF-SRT-EMB-LegalBERT) in Table~\ref{tab:leaderboard} are mixed. In the two ECtHR tasks and EUR-LEX, both models outperform the TFIDF-SVM baseline, a much simpler linear BoW model. Contrary, both models are outperformed by TFIDF-SVM in SCOTUS and LEDGAR. In UNFAIR-ToS, the three models perform overall on par. While the original word order is lost in all three models, TFIDF-SVM relies on $n$-grams up to 3 words long, which allows it to retain local word order in features that represent multi-word terms, like `civil rights' or `federal taxation' in the case of SCOTUS. We suspect that such multi-word terms are more important in SCOTUS and LEDGAR, which would explain the fact that TFIDF-SVM outperforms the other two BoW models in these tasks. Future work could add a TFIDF-SVM variant with only unigram features to check this hypothesis; there should be a large performance drop in the three tasks. One could also explore ways to use TF-IDF information about $n$-grams (not just unigrams) in the BoW variants of LegalBERT.

Switching to the aggregated results of the long document tasks of Table \ref{tab:avg_long_leaderboard}, we observe that both BoW variants of LegalBERT outperform TFIDF-SVM. Table~\ref{tab:avg_long_leaderboard} also shows that TFIDF-SRT-EMB-LegalBERT (which includes the TF-IDF embeddings layer) performs slightly better than TFIDF-SRT-LegalBERT 
in terms of \macrof (1 p.p.\ improvement), but there is almost no difference in  \microf, and the results of Table~\ref{tab:leaderboard} show no clear winner between the two methods across tasks.\vspace{2mm}

\begin{table}[t]
    \centering
    \resizebox{\columnwidth}{!}{
    \begin{tabular}{l|cc|}
          \bf Method & \microf & \macrof \\
         \hline
         \hline
         TFIDF-SVM                      &  69.5 & 58.1 \\
         \hline
        TFIDF-SRT-LegalBERT              &  \underline{73.7} & 65.2 \\
        TFIDF-SRT-EMB-LegalBERT           &  73.6 & \underline{66.1} \\
         \hline
         \hline
        LegalBERT  &  \underline{75.4} & \underline{68.1} \\
         TFIDF-EMB-LegalBERT         & 74.6 & 66.3 \\
         \hline
         \hline
         Longformer     &  73.9 & 66.6 \\
        Longformer-8192                &  74.4 & 66.1 \\
        Longformer-8192-PAR                &  74.4 & 66.8 \\
        \hline
    LegalLongformer                  &  76.1 & 68.6 \\
    LegalLongformer-8192               &  \bf \underline{76.5} & 68.4 \\
    LegalLongformer-8192-PAR           &  75.6 & \bf \underline{69.4} \\
    \end{tabular}
    }
    \vspace{-2mm}
    \caption{Test results aggregated (harmonic mean) over the long-document classification tasks (ECtHR Tasks A and B, SCOTUS) of LexGLUE. Best scores per group are \underline{underlined}, and best overall are in \textbf{bold}.}
    \label{tab:avg_long_leaderboard}
\end{table}

\noindent\textbf{LegalBERT variants that retain word order}: Table~\ref{tab:leaderboard} shows that adding the TF-IDF embeddings layer to LegalBERT (TF-IDF-EMB-LegalBERT), without word deduplication and retaining the original word order, leads to lower performance in 5 out of 6 tasks compared to the original LegalBERT; LEDGAR is the only exception, with small improvements. The aggregated results of Table~\ref{tab:avg_long_leaderboard} also show that TF-IDF-EMB-LegalBERT is worse than the original LegalBERT. We can only hypothesize that TF-IDF-EMB-LegalBERT is in most cases unable to learn how to use the additional information from the additive TF-IDF embeddings, which are added only during fine-tuning (they were not present during pre-training). This hypothesis is based on the positive (albeit small) impact of the TF-IDF embeddings layer on LEDGAR, the largest dataset with 60k training examples. All other datasets contain fewer than 10k training examples (Table~\ref{tab:datasets_summary}), with the exception of EUR-LEX (55k), which does not support our hypothesis. 

Given appropriate computing resources, one could further pre-train TFIDF-EMB-LegalBERT to help it learn how to exploit the newly introduced TF-IDF embeddings. The same applies to both BoW variants of LegalBERT, although in that case appropriate BoW pre-training objectives should be considered, since Masked Language Modeling (MLM) is not reasonable when the original word order is lost. Predicting the TF-IDF bucket id when masked, or predicting masked words given their TF-IDF bucket ids seem better alternatives.

\begin{table*}[t]
    \centering
    \resizebox{\textwidth}{!}{
    \begin{tabular}{l|c|cc|cc|cc|cc|cc}
          \multirow{2}{*}{\bf Method} & \multirow{2}{*}{\bf Params.} &  \multicolumn{2}{c|}{\bf ECtHR*} & \multicolumn{2}{c|}{\bf SCOTUS*} & \multicolumn{2}{c|}{\bf EUR-LEX} & \multicolumn{2}{c|}{\bf LEDGAR} & \multicolumn{2}{c}{\bf UNFAIR-ToS}\\
          & & Mem. & Time  & Mem. & Time & Mem. & Time & Mem. & Time & Mem. & Time \\
         \hline
         \hline
         \multicolumn{12}{c}{BoW models (word order lost)} \\
         \hline
         \hline
         TFIDF-SVM                      & 0.5M & 0.1 & .001 & 0.1 & .001 & 0.1 & .001 & 0.1 & .001 & 0.1 & .001 \\
         \hline
        TFIDF-SRT-LegalBERT              &  110M & 0.9 & .012 & 0.9 & .012 & 0.9 & .012 & 0.9 & .007  & 0.9 & .007 \\
        TFIDF-SRT-EMB-LegalBERT            &  110M & 0.9 & .012 & 0.9 & .012 & 0.9 & .012 & 0.9 & .007  & 0.9 & .007 \\
         \hline
         \hline
         \multicolumn{12}{c}{LegalBERT variants that retain word order} \\
         \hline
         \hline
        LegalBERT   &  110M & 1.3 & .014 & 1.3 & .014 & 1.9 & .012 & 1.9 & .007 & 1.9 & .007 \\
         TFIDF-EMB-LegalBERT        &  110M & 1.3 & .014 & 1.3 & .014 & 1.9 & .012 & 1.9 & .007 & 1.9 & .007 \\
        \hline
         \hline
         \multicolumn{12}{c}{Longformer variants (all retain word order)} \\
         \hline
         \hline
         LegalLongformer     &  148M & 1.7 & .164 & 1.7 & .164 & 1.3 & .033 & 1.3 & .033 & 1.3 & .033 \\
LegalLongformer-8192              &  151M & 2.2 & .318 & 2.2 & .318 & \multicolumn{6}{c}{(Not considered for short-document tasks.)} \\
LegalLongformer-8192-PAR               &  151M & 2.2 & .331 & 2.2 & .331 & \multicolumn{6}{c}{$>>$} \\
    \end{tabular}
    }
    \caption{Model parameters, memory footprint (GBs/sample), and inference time (sec/sample). In starred tasks, we use the hierarchical variant of LegalBERT. For ECtHR Tasks A and B, the information of this table is identical. 
    }
    \label{tab:efficiency_scores}
\end{table*}

\noindent\textbf{Longformer variants}:
Comparing the original Longformer with Longformer-8192, a variant capable of processing even longer documents, the results are mixed (Table~\ref{tab:leaderboard}) across the 3 long document classification tasks (ECtHR Tasks A and B, SCOTUS), i.e., \microf is improved at the expense of \macrof, or vice-versa. Aggregating the results (Table~\ref{tab:avg_long_leaderboard}), we observe the very same trade-off (+0.5 p.p.\ in \microf, -0.5 p.p.\ in \macrof). Considering the additional global tokens in Longformer-8192-PAR, we have comparable results in ECtHR tasks and improved results in SCOTUS, the dataset with the longest documents in LexGLUE (Table~\ref{tab:datasets_summary}). Aggregating the results (Table~\ref{tab:avg_long_leaderboard}), we observe that the extra global tokens do not improve \microf further (74.4), but lead to the best \macrof (66.8) of all the Longformer variants that have not been pre-trained on legal corpora.
Based on the aforementioned observations, we believe that the additional positional embeddings and adding more global tokens are in the right direction when seeking better long document performance with Longformer.  

Moving on to Longformer variants warm-started from LegalBERT, Table~\ref{tab:leaderboard} shows that LegalLongformer outperforms the 
original generic Longformer \cite{Longformer} in most cases, which highlights the importance of domain-specific models as already noted in the literature \cite{chalkidis-etal-2021-lexglue, Zheng2021}. We observe notable improvements in long document classification tasks (ECtHR A and B, SCOTUS), with approx.\ +2.0 p.p.\ in both \microf and \macrof in the aggregated results of Table~\ref{tab:avg_long_leaderboard}. These results are impressive considering that LegalLongformer was warm-started from LegalBERT, but no additional pre-training was conducted; hence several parameters of the model (e.g., additional positional embeddings and global attention matrices) may be far from optimal. By contrast, the original Longformer was warm-started from RoBERTa and was pre-trained for 64k additional steps on generic long documents.

Considering the last two variants of LegalLongformer (-8192, -8192-PAR), the results are mixed (trade-off between \microf and \macrof in Table~\ref{tab:leaderboard}, as with the generic  Longformer) and share the best aggregated results across all examined methods in long document classifications tasks (Table~\ref{tab:avg_long_leaderboard}).

Based on the above, we believe that the proposed extensions (warm-start from a legally pre-trained model, additional positional embeddings, additional global tokens) are in the right direction, already producing better results compared to the generic Longformer, and state-of-the-art results in several LexGLUE tasks (ECtHR A\&B and LEDGAR).
Given appropriate resources, one could further pre-train LegalLongformer-8192-PAR  for a limited number of steps (e.g., 64k) on long legal documents (e.g., the training subsets of ECtHR, and SCOTUS) to optimize the newly introduced parameters and  expect further improvements.

\subsection{Efficiency considerations}

In Table~\ref{tab:efficiency_scores}, we present important information with respect to efficiency. As expected,  TFIDF-SVM has the fewest parameters (200$\times$ fewer than LegalBERT variants) and is substantially faster and less memory-intensive compared to all other neural methods, while achieving state-of-the-art results in two tasks (SCOTUS and EUR-LEX, Table~\ref{tab:leaderboard}). 

Our proposed BoW variants of LegalBERT are substantially less memory intensive; approx.\ 25\% less GPU memory across the long document classification tasks (starred), and approx.\ 50\% less GPU memory across others with much shortened texts compared to LegalBERT. The TF-IDF embeddings do not affect memory or inference time (storing and looking up TF-IDF embeddings are negligible). 

Considering LegalLongformer, we observe an approx.\ 50\% increase in the number of parameters and approx.\ 25\% increase in GPU memory. With respect to inference time, there is a 10$\times$  increase compared to LegalBERT models in long document processing tasks, and larger in the other tasks with much shorter documents, which makes hierarchical Transformers a faster alternative. 

Moving to the extensions of LegalLongformer that are able to encode longer documents (LegalLongformer8192) and use extra global tokens (LegalLongformer-8192-PAR), there is an approx. 30\% increase in GPU memory compared to the standard Longformer (encoding up to 4,096 sub-words), and 2$\times$ increase in inference time. In other words, there is no free lunch when seeking performance improvements.

\section{Conclusions and Future Work}

Concluding, we presented BoW variants of LegalBERT, which remove duplicate words and consider TF-IDF scores by reordering the remaining words and/or by employing a TF-IDF embedding layer. These variants are more efficient than the original LegalBERT and still overall outperform a TF-IDF-based SVM in long legal document classification.  

We also modified Longformer to handle even longer texts (up to 8,192 sub-words), use additional global tokens, and also showed the positive effect of warm-starting it from LegalBERT. Unlike the BoW models, this is a resource-intensive direction, with substantial improvements compared to the original Longformer (up to 4,096 sub-words, a single global token, warm-started from RoBERTa) in long legal document classification. The new LegalLongformer (and its variants) are the new state of the art in the long document tasks of LexGLUE.

In future work, we would like to further pre-train the proposed BoW variants of LegalBERT, LegalLongformer, and variants on legal corpora, to help them better optimize the newly introduced modifications (e.g., TF-IDF embeddings, additional positional embeddings, updated attention scheme with additional global tokens). We would also like to experiment with long documents from other domains (e.g., long business documents). 

\section*{Acknowledgments}
This research has been co‐funded by the European Regional Development Fund of the European Union and Greek national funds through the Operational Program Competitiveness, Entrepreneurship and Innovation, under the call RESEARCH – CREATE – INNOVATE (Τ2ΕΔΚ-03849).  This work is also partly funded by the Innovation Fund Denmark (IFD)\footnote{\url{https://innovationsfonden.dk/en}} under File No.\ 0175-00011A.

\bibliography{anthology,acl2020}
\bibliographystyle{acl_natbib}
\appendix

\section{Additional Material}

\begin{figure*}
\centering
\includegraphics[width=0.7\linewidth]{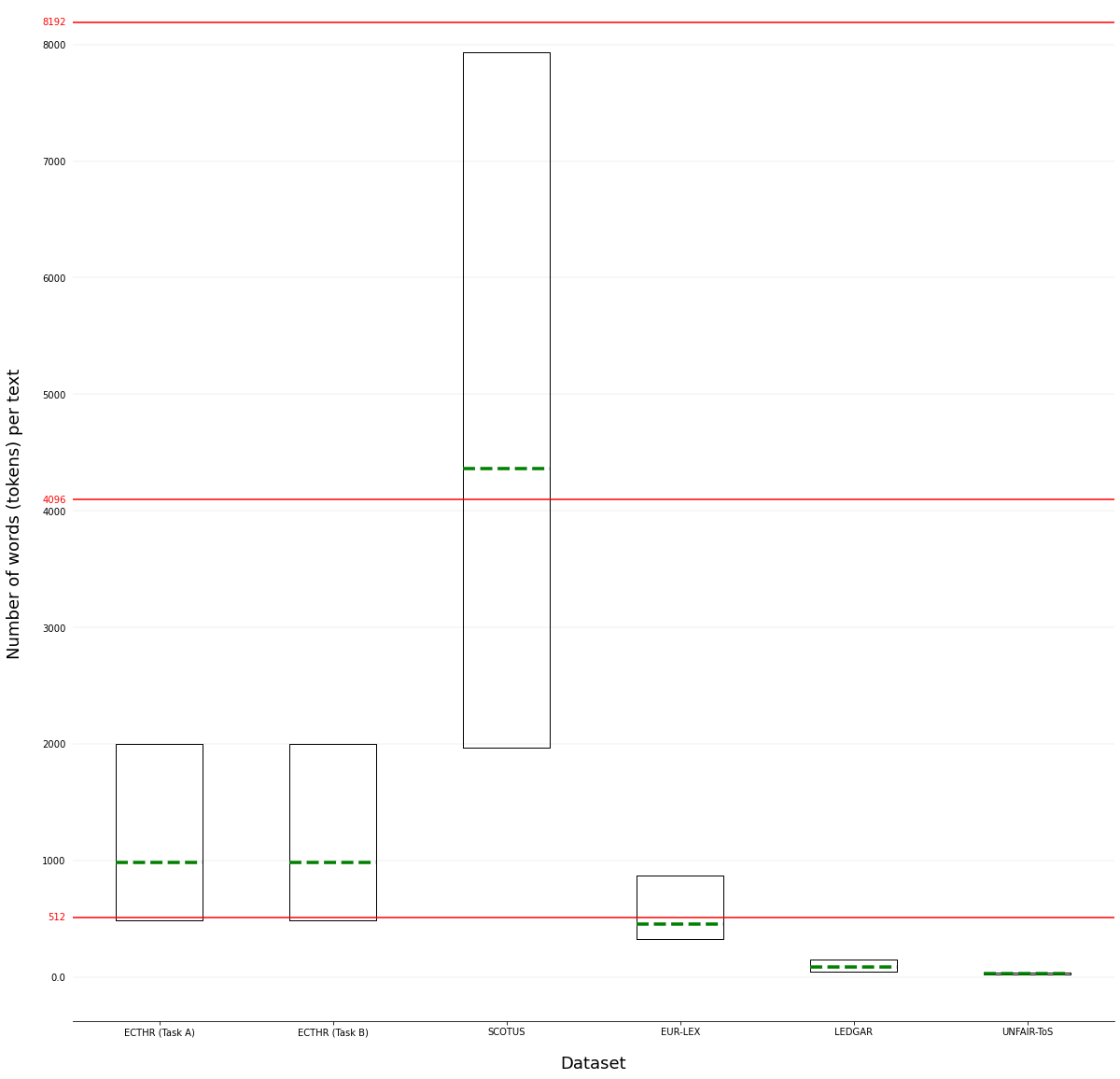}
\caption{Average text length (in sub-words) across LexGLUE datasets \underline{before} 
word deduplication.}
\label{tab:boxplot1}
\end{figure*}

\begin{figure*}
\centering
\includegraphics[width=0.7\linewidth]{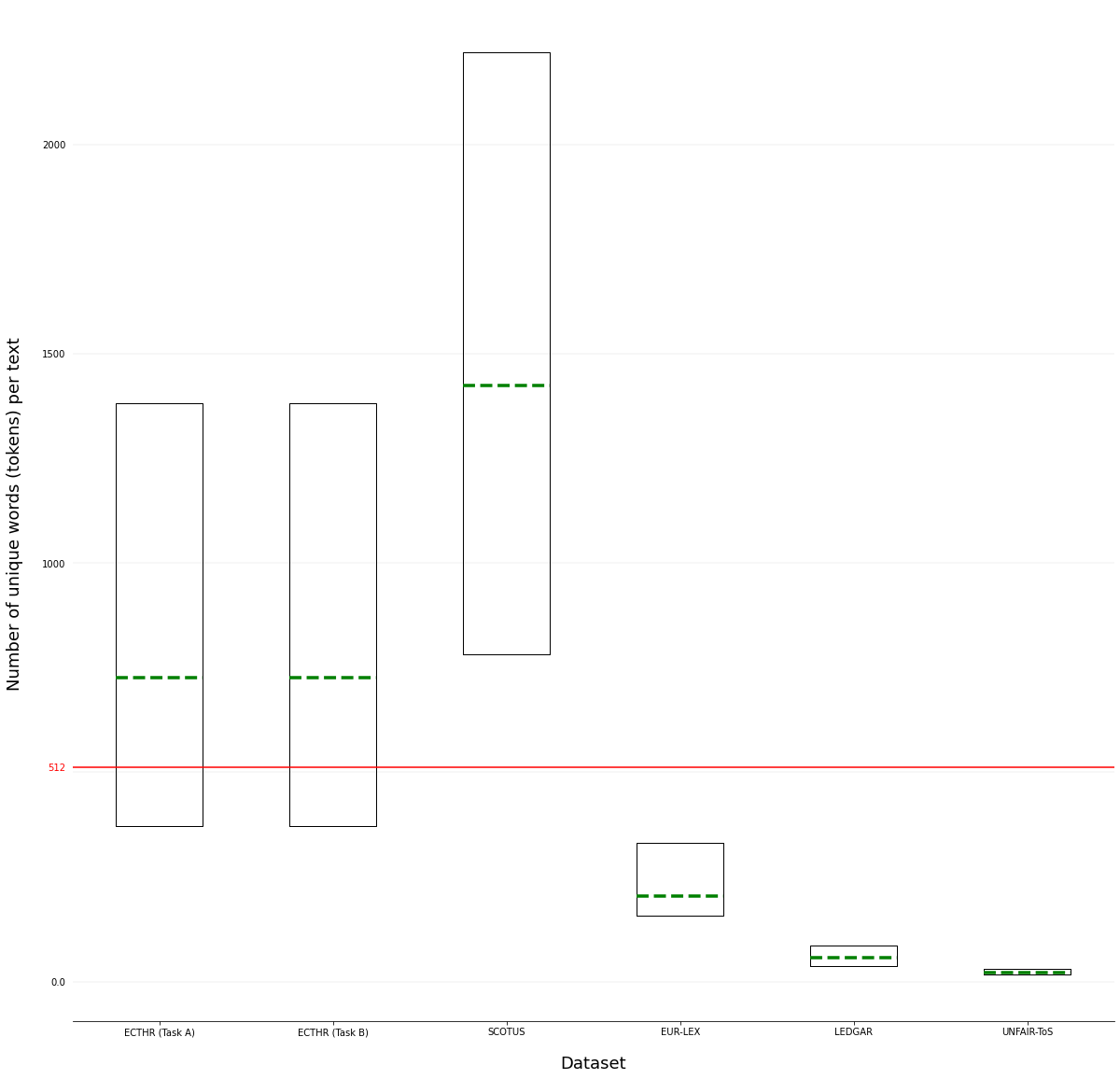}
\caption{Average text length (in sub-words) across LexGLUE datasets \underline{after} 
word deduplication.}
\label{tab:boxplot2}
\end{figure*}

In Figures~\ref{tab:boxplot1}--\ref{tab:boxplot2}, we show boxplots of the average text length (in sub-words) across LexGLUE datasets before and after word deduplication.

\end{document}